\documentclass{article}

\pdfoutput=1

\usepackage{arxiv}

\usepackage{physics}
\usepackage{amsmath}
\usepackage{amsfonts}       

\usepackage{setspace}
\usepackage{nicefrac}
\usepackage{array}
\usepackage{caption}
\usepackage{subcaption}
\usepackage[binary-units]{siunitx}
\usepackage{wrapfig}
\usepackage{bm}
\usepackage{scalerel}
\usepackage{tikz}
\usetikzlibrary{svg.path}

\usepackage[parfill]{parskip}
\usepackage[utf8]{inputenc} 
\usepackage[T1]{fontenc}    
\usepackage{hyperref}       
\usepackage{url}            
\usepackage{booktabs}       
\usepackage{nicefrac}       
\usepackage{microtype}      
\usepackage{lipsum}		
\usepackage{graphicx}
\usepackage{natbib}
\usepackage{doi}

\newcommand{\bs}[1]{\boldsymbol{#1}}
\newcolumntype{L}{>{\centering\arraybackslash}m{6cm}}

\title{A Convolutional Neural Network-based Approach to Field Reconstruction}

\author{ \href{https://orcid.org/0000-0002-7941-3523}{\includegraphics[scale=0.06]{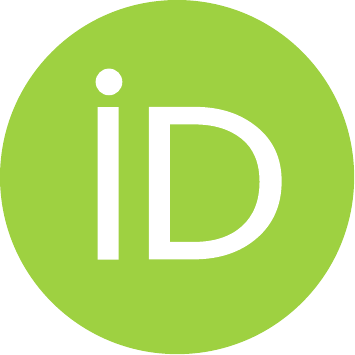}\hspace{1mm}Roberto Ponciroli} \\
	Nuclear Science and Engineering Division\\
	Argonne National Laboratory\\
	Lemont, IL 60439 \\
	\texttt{rponciroli@anl.gov} \\
	\And
	\href{https://orcid.org/0000-0002-6971-7769}{\includegraphics[scale=0.06]{orcid.pdf}\hspace{1mm}Andrea Rovinelli} \\
	Applied Materials Division\\
	Argonne National Laboratory\\
	Lemont, IL 60439 \\
	\texttt{arovinelli@anl.gov} \\
	\And
	\href{https://orcid.org/0000-0002-1580-5560}{\includegraphics[scale=0.06]{orcid.pdf}\hspace{1mm}Lander Ibarra} \\
	Nuclear Science and Engineering Division\\
	Argonne National Laboratory\\
	Lemont, IL 60439 \\
	\texttt{libarra@anl.gov} \\
}

\renewcommand{\shorttitle}{CNN-based Approach to Field Reconstruction}

\hypersetup{
pdftitle={A Convolutional Neural Network-based Approach to Field Reconstruction},
pdfsubject={CC Field Reconstruction},
pdfauthor={R. Ponciroli, A. Rovinelli, L. Ibarra},
pdfkeywords={Convolutional Neural Network, Data-driven approach, Diagnostics, Field Reconstruction},
}

\begin{document}
\maketitle

\begin{abstract}
	This work has been submitted to the IEEE for possible publication. Copyright may be transferred without notice, after which this version may no longer be accessible.

 \vspace{\baselineskip}

	In many applications, the spatial distribution of a field needs to be carefully monitored to detect spikes, discontinuities or dangerous heterogeneities, but invasive monitoring approaches cannot be used. Besides, technical specifications about the process might not be available by preventing the adoption of an accurate model of the system. In this work, a physics-informed, data-driven algorithm that allows addressing these requirements is presented. The approach is based on the implementation of a boundary element method (BEM)-scheme within a convolutional neural network. Thanks to the capability of representing any continuous mathematical function with a reduced number of parameters, the network allows predicting the field value in any point of the domain, given the boundary conditions and few measurements within the domain. The proposed approach was applied to reconstruct a field described by the Helmholtz equation over a three-dimensional domain. A sensitivity analysis was also performed by investigating different physical conditions and different network configurations. Since the only assumption is the applicability of BEM, the current approach can be applied to the monitoring of a wide range of processes, from the localization of the source of pollutant within a water reservoir to the monitoring of the neutron flux in a nuclear reactor.
\end{abstract}


\section{Introduction} 
\label{sec:intro}
In many engineering applications, the spatial distribution of a field (scalar or vector) needs to be carefully monitored to detect local spikes, discontinuities, or dangerous heterogeneities. This can be achieved through a matrix of sensors sufficiently dense to resolve the spatial characteristics of the field. Sometimes, the nature itself of the monitored process does not allow placing as many sensors as needed inside the domain to resolve the spatial characteristics of the field. At the same time, the restrictions on the boundary are in general not as stringent. Mathematically, this diagnostics problem can be formulated as a boundary value (BV) problem as described by \cite{russell1975numerical}. A BV problem is a problem of determining a solution to a differential equation subject to conditions on the unknown function specified at two or more values of the independent variable (boundary conditions). One of the most used techniques to solve this class of problems is the boundary element method (BEM) which has been applied with success for a variety of research fields and applications. A few representative demonstrations of this methodology can be founf in  \cite{blobner1999transient} \cite{beer2008boundary} and \cite{godin1996kirchhoff}. Unlike finite domain methods, the BEM formulates the BV problems as a system of boundary integral equations. This methodology requires discretizing only the domain surface instead of the entire domain, and mimics the physical limitations of the diagnostics problem (the field within the domain can be reconstructed starting from the measurements collected on the domain boundary).

Among the diagnostics applications of BEM, the acoustic holography is a powerful, non-invasive technique for the identification and localization of vibratory sources as discussed by \cite{langrenne2007boundary}. By placing a set of pressure sensors at points located on a surface in proximity to an acoustically radiating object, it allows reconstructing the velocity vector, the acoustic pressure, and the acoustic intensity map in a three-dimensional space per \cite{Hayek2008}. The key assumption that allows the application of this technique is the knowledge of the physics of the process (acoustic field), i.e., the Green’s function of the differential operator needs to be known a-priori. In the perspective of extending this approach to the monitoring of other physical processes, sometimes very limited information is available about the studied system and/or the parameters characterizing the governing differential equation. In this work, a convolutional neural network (CNN) similar to the description in \cite{gopika2020single} is proposed to retrieve the Green’s function and to reconstruct the spatial distribution of diffusive and/or advective fields over a domain of arbitrary geometry. Thanks to the capability of representing any continuous mathematical function with a relatively reduced number of parameters, the designed network allows predicting the value of the field in any point within the domain starting from (1) the accurate characterization of the field on the domain boundaries (simultaneous knowledge of the Dirichlet and the Neumann boundary conditions), and (2) the evaluation of the field in few locations within the domain. In the last few years, deep neural networks have attracted attention for data modeling and solving differential equations. Through the training, the weights and biases are optimized so that the network outputs the closest approximation of the solution of the equations. In the publication from \cite{owhadi2015bayesian}, the general framework of solving differential equations as a learning problem was proposed, and in the paper \cite{xu2019neural}, the possibility of directly using preexisting finite discretization schemes within the loss function was examined. From this standpoint, because of the implementation of the BEM within a data-driven scheme, the proposed framework can be referred to as physics-informed neural network (PINN) as described by \cite{raissi}.

The paper is organized as follows. In Section \ref{sec:fiedlchar}, the formulation of the field reconstruction as a BV problem is presented, and the traditional BEM approach is briefly described. In Section \ref{sec:CNNalg}, the designed neural network for the field reconstruction is presented. In Section \ref{sec:testcase}, the performance of the network is assessed on a reference test-case. The training and the testing procedures of the network are presented in great detail in Sections \ref{sec:training} and \ref{sec:testing}, respectively. To evaluate the versatility of the proposed algorithm, a sensitivity analysis was performed by investigating different physical conditions and by adopting different network configurations (Section \ref{sec:method}). Finally, the main conclusions are drawn (Section \ref{sec:conclusion}).
\section{Formulation of the Field reconstruction problem}
\label{sec:fiedlchar}
\subsection{Boundary Value problem} \label{sec:bvp}
A BV problem for a differential equation consists of finding a solution of the given differential equation subject to a set of boundary conditions (Eq.(\ref{eq:eq1})).
\begin{equation}
\begin{split}
\left(L u\right)\left(\bs{r} \right)=f\left(\bs{r} \right)
\textrm{ subject to } u(\bs{r'}) = u_0 \textrm{ and } \\
\frac{\partial u}{\partial \bs{n}}(\bs{r'}) = \nabla u(\bs{r'}) \cdot \bs{n} = u'_0 \textrm{ with } \bs{r} \in \Omega, \bs{r'} \in \Gamma
\end{split}
\label{eq:eq1}
\end{equation}

where $\bs{r}$ represents a position vector of a location where the solution of $u \left(\bs{r}\right)$ needs to be evaluated, $\bs{r'}$ represents the position vectors of the points on the boundary ($\Gamma$) where conditions are imposed, $L$ is the differential operator of the equation over the domain ($\Omega$), and $f\left(\bs{r}\right)$ is the source term.\\
The main focus of this work is the reconstruction of scalar fields described by the Helmholtz equation over domains containing media with homogeneous characteristics. This equation describes the spatial distribution of the physical quantities, such as temperature, pressure, etc., characterizing a wide variety of diffusive and/or advective phenomena. In Eq.(\ref{eq:eq2}), the differential operator for the Helmholtz equation is reported.

\begin{equation}
L u = \left (\nabla^2 + k^2 \right) u
\label{eq:eq2}
\end{equation}

where $k$ represents the wavenumber. \\
The solution to Eq.(\ref{eq:eq2}) must also be satisfied on the boundary for any $\bs{r'}$ position vector.

\begin{equation}
\left(Bu\right)\left(\bs{r'}\right) = \Phi\left(\bs{r'}\right)
\label{eq:eq3}
\end{equation}

where $B$ represents the differential operator of the Helmholtz equation, and $\Phi\left(\bs{r'}\right)$ represents the source term, both defined on the boundary.

\subsection{Boundary Element Method} \label{sec:bve}
A powerful tool to evaluate the solution of the scalar time-independent Helmholtz equation is the Kirchhoff–Helmholtz integral equation described by \cite{godin1996kirchhoff}. By applying Green’s third identity, the initial differential equation can be transformed into an integral equation, and the complex amplitude of the field at a point within the domain can be related to the complex amplitude of the field on the enclosing surface (Eq.(\ref{eq:eq4})).

\begin{equation}
  u(\bs{r})\eta(\bs{r})=\int_{\Gamma} \left(u(\bs{r}') \frac{\partial G(\bs{r},\bs{r}')}{\partial \bs{n}}-G(\bs{r},\bs{r}')\frac{\partial u(\bs{r}')}{\partial \bs{n}} \right) \dd \Gamma - \int_{\Omega} G(\bs{r},\bs{r}') f(\bs{r}) \dd \Omega \textrm{ with } \bs{r} \in \Omega \subset \mathbb{R}^\textrm{d}
\label{eq:eq4}
\end{equation}

where $G$ represents the Green’s function (or fundamental solution) corresponding to the differential operator ($L$), $d=2,3$ is a parameter accounting for the dimensionality of the domain, and $\eta(\bs{r})$ is a solid angle coefficient assuming the values reported in Eq.(\ref{eq:c_P}).

\begin{equation}
   \eta(\bs{r}) =
    \begin{cases}
        1 & \mbox{ if } \bs{r} \in \Omega \\
        \frac{1}{2} & \mbox{ if } \bs{r} \in \Gamma \mbox{ and } \Gamma \mbox{ smooth at } \bs{r} \\
        \frac{\mbox{inner solid angle}}{4\pi} & \mbox{ if } \bs{r} \in \Gamma \mbox{ and } \Gamma \mbox{ not smooth at } \bs{r}
    \end{cases}
   \label{eq:c_P}
\end{equation}

The boundary element method (BEM) is a numerical computational method of solving linear partial differential equations if these can be formulated as integral equations similarly to the description from \cite{betcke2019boundary}. The BEM approach is based on the discretization of the Kirchhoff–Helmholtz integral equation with no source (Eq.(\ref{eq:BEM})).
\begin{equation}
    u(\bs{r}_i)\eta(\bs{r}_i)=\sum_{j=1}^{N_c}\left[\int_{\Gamma_j} u(\bs{r'}_j) \frac{\partial G(\bs{r}_i, \bs{r'}_j)}{\partial \bs{n}_j} \,d\Gamma\right]- \sum_{j=1}^{N_c}\left[\int_{\Gamma_j} G(\bs{r}_i, \bs{r'}_j) \frac{\partial u(\bs{r'}_j)}{\partial\bs{n}_j} \,d\Gamma\right]
    \textrm{ with } \bs{r}_i \in \Omega, \bs{r'}_j \in \Gamma_j \subset \Gamma
\label{eq:BEM}
\end{equation}
where $N_{c}$ is the number of elements ($\Gamma_j$) that constitute the boundary of the domain.

The numerical solution of the Kirchhoff–Helmholtz integral equation requires the knowledge of the Green’s function and its normal derivative at different locations on the boundary. These points are called \emph{collocation points}, and satisfy the solution of Eq.(\ref{eq:eq3}). Similarly to the finite element method, basis functions are employed to calculate the field distribution on the boundary. For the purpose of this study, mixed Dirichlet-Neumann boundary conditions are imposed on these collocation points. A field distribution can be approximated at a selected number of points on the boundary of a finite-dimensional space. Once the solution is obtained on the boundary, the integral equation is used to calculate the field distribution. The solution within the domain is not known unless the exact solution is computed for specific points. The locations where the interior field is evaluated are called \emph{interior points} herein.

The main advantage of the BEM is that the dimensional reduction of the problem. Because the Kirchhoff–Helmholtz integral equation is first solved on the boundary of the domain, spatial dimensions are reduced by one with respect to other numerical methods (e.g., finite element or finite volume  methods). Besides, for the same level of accuracy, the BEM uses a lesser number of nodes and elements. These characteristics are extremely attractive when applied to monitoring and domain reconstruction  where fast algorithms and solvers are needed.

\section{Design of a Convolutional Neural Network for field reconstruction} \label{sec:CNNalg}
\subsection{Diagnostics requirements} \label{sec:diagnostic}
The main focus of this work is the presentation of a tool that address the requirements of many monitoring and diagnostics applications. Let us assume that the distribution of a three-dimensional field needs to be reconstructed, and that the number of sensors that can be placed within the domain is too small to ensure the desired spatial resolution. At the same time, the restrictions on the boundary are not so stringent, and a much larger sensor set is available. Let us also assume that the only piece of information about the physics of the process is that it is described by the Helmholtz equation. To address all these requirements, we tackled the monitoring problem as a BV problem that can be solved by adopting the Kirchhoff–Helmholtz integral equation (BEM approach). The key ingredient is constituted by the Green’s function corresponding to the differential operator of the equation governing the process. Given the very limited amount of information, this function is not known a-priori. The proposed algorithm allows retrieving it from the available sensor readings through an automated, data-driven learning process. Thanks to the capability of representing any continuous mathematical function with a relatively reduced number of parameters, a neural network-based approach is proposed. Given that the nature of the Kirchhoff–Helmholtz equation, a CNN would represent the most suitable scheme. CNNs are a category of neural networks that have proven very effective in areas such as image recognition and classification as demonstrated by \cite{kharazmi2019variational}. The convolution is a linear operation that extracts features from the input image by preserving the spatial relationship between the pixels. It involves the multiplication of the filter with the array of input data (the pixel values), i.e., it is an element-wise multiplication which is then summed, always resulting in a single value. This capability is crucial to calculate the integrals that appear in the Kirchhoff-Helmholtz integral equation. In Table \ref{tab:table1}, the features of the traditional BEM and the features of the proposed CNN-based BEM algorithm are reported.

\begin{table*}[ht!]
\newcolumntype{Q}{>{\centering\arraybackslash}m{6cm}}
\centering
\small
\captionsetup{width=.75\textwidth}
\caption{Comparison  of  the  features of the  traditional BEM with respect to the proposed CNN-based approach.}
\begin{tabular}{QQ}
\hline
\textbf{Traditional BEM approach} & \textbf{CNN-based BEM approach} \\ \hline
The Green’s function is known. & The Green’s function is not known a-priori. It is evaluated through a supervised learning process. \\
 & \\
The values of either $u(\bs{r'})$ or $\partial u(\bs{r'}) / \partial \bs{n}$ on the domain boundary need to be provided. & The values of both $u(\bs{r'})$ and $\partial u(\bs{r'}) / \partial \bs{n}$ on the domain boundary need to be provided.\\
 & \\
Field can be evaluated in any point of the domain. & Field values in few locations need to be initially provided. Once the network is trained, the field can be evaluated in any point of the domain. \\ \hline
\end{tabular}
\label{tab:table1}
\end{table*}

\subsection{Features of the designed network for solving the BV problem} \label{sec:CNNBVP}
\begin{figure*}[ht!]
      \centering
      \includegraphics[width=0.7\textwidth]{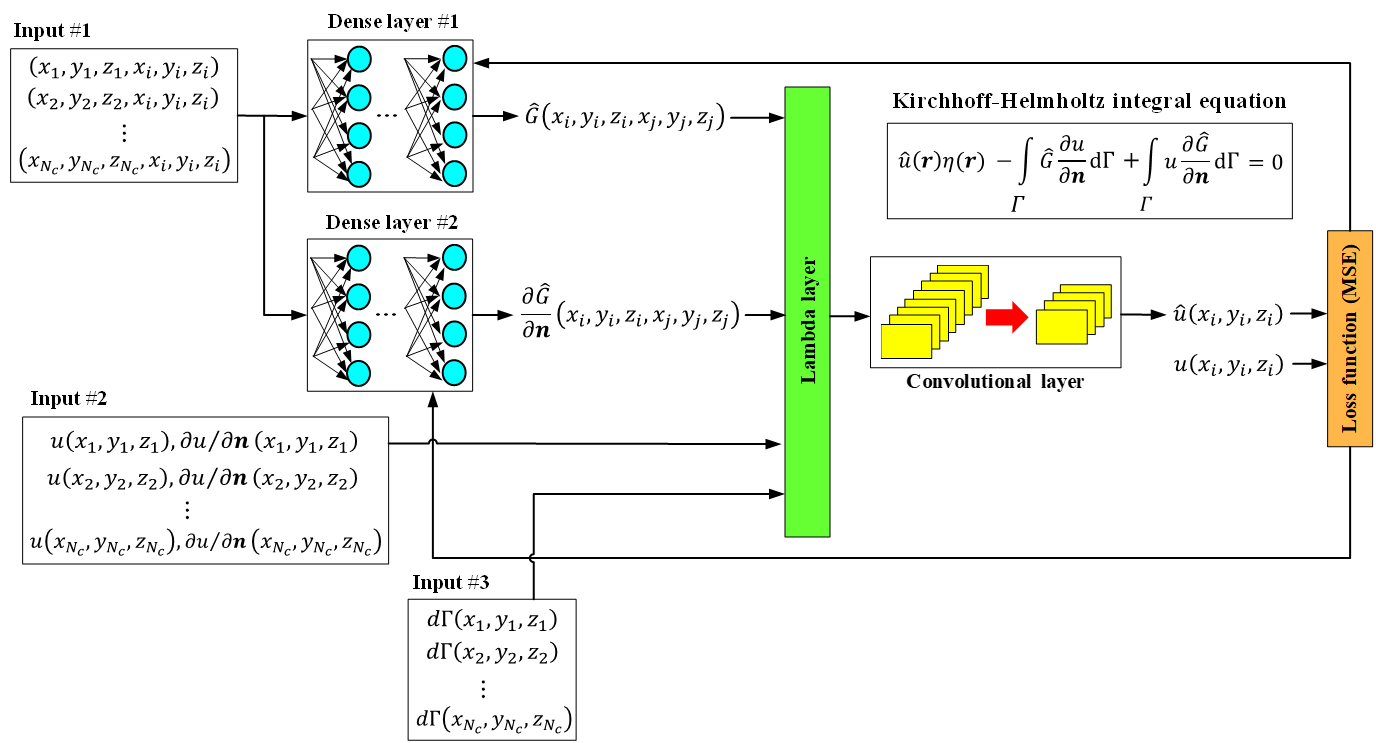}
      \caption{Graphical representation of the developed CNN for field reconstruction.}
      \label{fig:CNN}
\end{figure*}

In this Section, the main features of the designed CNN are reported. In Figure \ref{fig:CNN}, the graphical representation of the network is shown. In Table \ref{tab:table2}, details about the provided datasets and their shapes are reported.
\begin{itemize}
\item The first input ("Input \#1") provides the coordinates of the collocation points ($x_j$, $y_j$, $z_j$), where the boundary conditions are evaluated, and the interior points ($x_i$, $y_i$, $z_i$), where the field is measured.
\item The second input ("Input \#2") provides the values of Dirichlet $(u(x_j,y_j,z_j))$ and the Neumann ($\partial u(x_j,y_j,z_j) / \partial \bs{n} (x_j,y_j,z_j)$) boundary conditions (BCs).
\item The third input ("Input \#3") provides the size of the boundary element in correspondence of the collocation points ($\textrm{d}\Gamma (x_j,y_j,z_j)$).
\item Two sets of 3 dense layers ("Dense layer \#1" and "Dense layer \#2") are adopted to reconstruct the Green’s function corresponding to the differential operator of the Helmholtz equation and its normal derivative, $\hat{G}(x_i,y_i,z_i,x_j,y_j,z_j)$ and $\partial \hat{G} / \partial \bs{n} (x_i,y_i,z_i,x_j,y_j,z_j)$, respectively. Each set of layers processes the merged coordinates of interior and collocation points contained in "Input \#1", and returns linear combinations of these coordinates that, together with the imposed boundary conditions, will be used to predict the value of the field at the interior points. As stressed in Section \ref{sec:bve}, BEM can only be applied to problems for which the Green's functions are available. Usually, they can be calculated only for linear partial differential equations with constant or piecewise constant coefficients. In other cases, their numerical evaluation requires costly, numerical techniques like the Prony's method or the complex image method from \cite{Green} and \cite{Green2}. In this work, these functions are inferred through the supervised learning process. Similar approaches aiming at extracting physical parameters or obtaining the numerical approximation of differential operators through explicit embedding of the governing equations were recently proposed. In \cite{Lu}, an unsupervised learning technique using variational autoencoders to extract physical parameters from noisy spatiotemporal data was proposed.
\item A lambda layer allows calculating the integrand of the Kirchhoff-Helmholtz integral equation. 
\item A convolutional layer is used to integrate the output of the lambda layer over the domain borders. Traditionally, convolutional layers are placed at the beginning of the CNNs to reduce images into a form that is easier to process, without losing important features which are critical for generating accurate predictions. The current application of the convolutional layer is quite different. Unlike dense layers, this layer is not involved in the learning process. Once provided the inputs, the convolutional layer is expected to calculate the integrals that are involved in the Kirchhoff-Helmholtz equation. This is the reason why this layer constitutes the last step of the network, and the corresponding weights/biases are kept constant during the training.
\item The convolutional layer outputs the estimated value of the field at the interior points that are selected for the training/testing. The learning process stems from the minimization of the loss function that in this case is the mean square error (MSE) between the estimated value ($\hat{u}(x_i,y_i,z_i)$) and the measured value ($u(x_i,y_i,z_i)$) of the field evaluated at the interior points (Eq.(\ref{eq:MSE})). Starting from the value of the loss function, the weights of the linear combinations representing the Green's function and its normal derivative are updated through back-propagation method described by \cite{goodfellow2016deep}.

\begin{equation}
\label{eq:MSE}
e=\frac{1}{N_P} \sqrt{\sum_{i=1}^{N_P} \left(\hat{u}(x_i,y_i,z_i)-u(x_i,y_i,z_i) \right)^2}
\end{equation}
\end{itemize}

\begin{table*}[ht!]
\newcolumntype{Q}{>{\centering\arraybackslash}m{3cm}}
\centering
\small
\captionsetup{width=.5\textwidth}
\caption{Description of collocation points, interior points and inputs according to the dimensionality of the domain.}
\begin{tabular}{QQQQ}
\hline
\textbf{Variable} & \textbf{Description}  & \textbf{2D} & \textbf{3D}  \\\hline
$\bs{r'}$ & Collocation points coordinate, $j=1,2,...N_C$ & $(x_j,y_j)$ & $(x_j,y_j,z_j)$  \\  &&& \\
$\bs{r}$ & Interior points coordinate, $i=1,2,...N_P$ & $(x_i,y_i)$ & $(x_i,y_i,z_i)$  \\  &&& \\
\textbf{Input \#1} & Coordinates of Collocation points  and  Interior points & $\textrm{size}(N_P,N_C,4)$ & $\textrm{size}(N_P,N_C,6)$  \\  &&& \\
\textbf{Input \#2} & Dirichlet/Neumann BCs evaluated in correspondence of Collocation points & $\textrm{size}(N_P,N_C,2)$ & $\textrm{size}(N_P,N_C,2)$  \\  &&& \\
\textbf{Input \#3} & Sizes of the boundary elements in correspondence of Collocation points & $\textrm{size}(N_P,N_C,1)$ & $\textrm{size}(N_P,N_C,1)$
\\ \hline
\end{tabular}
\label{tab:table2}
\end{table*}

The described CNN was implemented into the Keras deep learning framework which is described in \cite{chollet2015keras}. Keras is an open-source software library that provides a Python interface for artificial neural networks and runs on top of Tensor-Flow library outlined in \cite{tensorflow}. Being a high-level neural networks API (application programming interface), it allows for easy and fast prototyping. To optimize the weights of the dense layer during the training of the network, the Adaptive Moment Estimation (Adam) method described in \cite{kingma2014adam} was used. It is a stochastic gradient descent algorithm based on adaptive estimates of lower-order moments per \cite{ruder2016overview}. Adam is straightforward to implement, it is computationally efficient, it has little memory requirements and it is well suited for problems that are large in terms of data and/or parameters. As important as the optimization algorithm, the choice of activation function in the output layer will determine the accuracy of the network predictions. Traditionally, a Rectified Linear Unit (ReLU) function, i.e., an element-wise operation that replaces negative pixel values in the feature map by zero, is used to introduce non-linearity in the CNN. In this case, the ReLU function cannot be adopted since all the negative input values would be turned into zero by affecting the resulting graph (negative values would not be mapped appropriately). The hyperbolic tangent was then adopted as activation function for the dense layers. With respect to ReLU, negative inputs will be mapped strongly negative, and the zero inputs will be mapped near zero. In addition, the hyperbolic tangent is differentiable and monotonic.

\section{Assessment of the Performance of the designed network}
\label{sec:testcase}
\begin{figure}[htb!]
      \centering
      \includegraphics[width=0.5\textwidth]{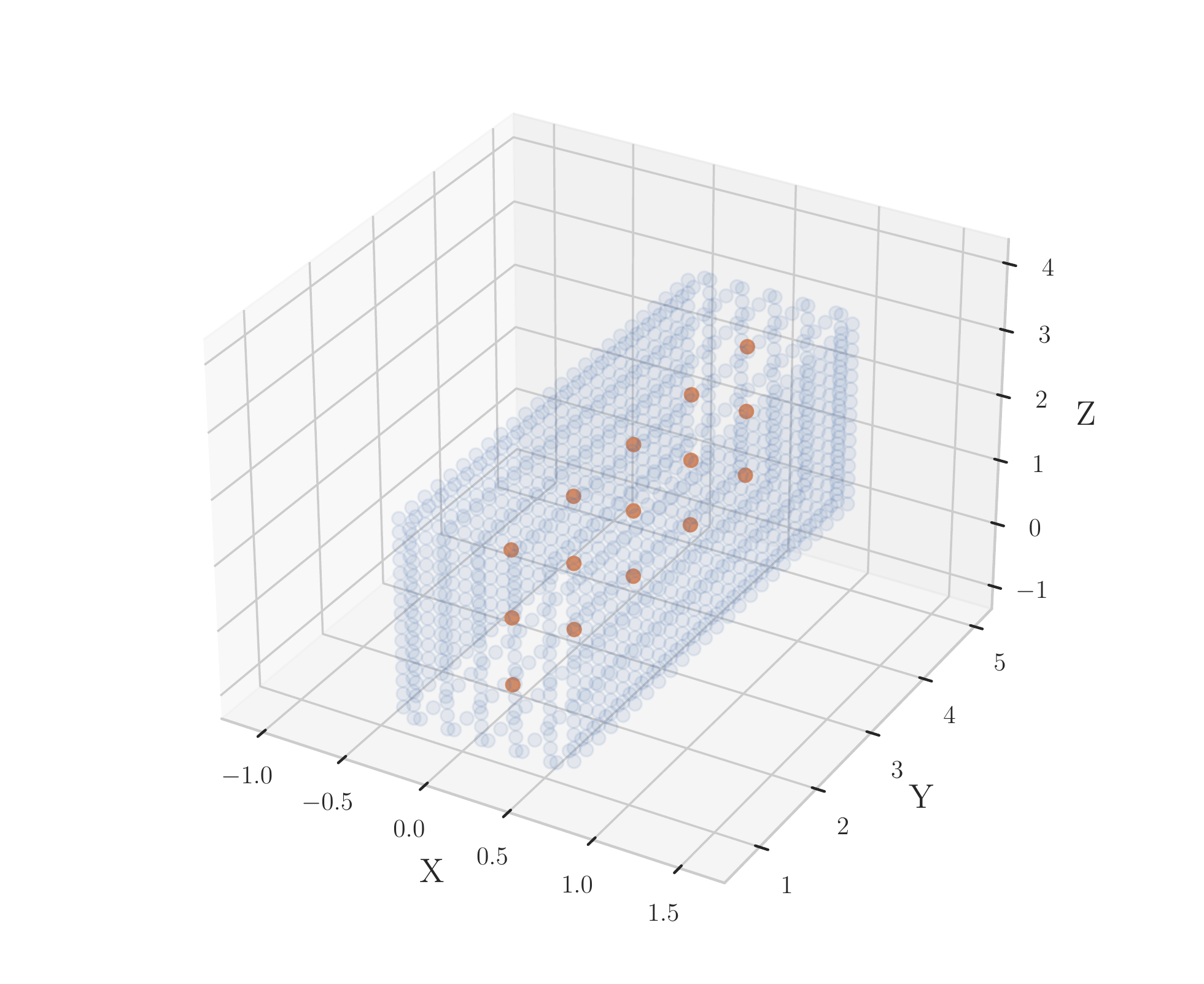}
      \caption{Representation of the 3D domain adopted as test-case.}
      \label{fig:InteriorPoints}
\end{figure}

To assess the performance of the proposed algorithm, a test-case was developed. The designed CNN was used to reconstruct the spatial distribution of the solution of the BV problem described by Eqs.(\ref{eq:eq5})(\ref{eq:eq6})(\ref{eq:eq7}) within a rectangular parallelepiped (Figure \ref{fig:InteriorPoints}).
\begin{equation}
\nabla^{2} u + u = 0
\label{eq:eq5}
\end{equation} 
\begin{equation}
u = 1\; \textrm{ on } \{\Gamma^{x+}\cup\Gamma^{z-}\}
\label{eq:eq6}
\end{equation}
\begin{equation}
\frac{\partial u}{\partial \bs{n}} = 1\; \textrm{ on } \;\Gamma - \{\Gamma^{x+}\cup\Gamma^{z-}\}
\label{eq:eq7}
\end{equation}

\subsection{Generation of training and testing datasets}
\label{sec:training}
The first step consists in defining the data sets containing the coordinates of the collocation points and the interior points that will be merged to constitute "Input \#1" (Figure \ref{fig:CNN}). The collocation points corresponds to the locations where the boundary conditions are available, "Input \#2". As mentioned before, in the foreseen applications of this algorithm, there are no specific restrictions to the number of sensors that can be placed on the domain boundary. For this reason, locations on the external surface were selected with steps equal to $\Delta x=0.1$ in all directions. Overall, the set of collocation points is constituted by \SI{4,600} elements. In Figure \ref{fig:InteriorPoints}, they are represented by blue circles.\\
The interior points assume different meanings in training and testing phases. During training, they represent the locations where the field can be measured, on the boundary or within the domain. These latter sensor readings are crucial to optimize the weights and the biases of the linear combinations used to approximate the Green’s function and its normal derivative. In Figure \ref{fig:InteriorPoints}, they are represented by \SI{15} red dots. During testing, the interior points represent the locations where the value of the field will be predicted. To obtain a high-resolution reconstruction, $15,000$ equally-spaced interior points ($n_x\cdot n_y\cdot n_z=10 \cdot 50 \cdot 30$) were used.\\
The second step consists in calculating the field values at the interior points for both training and testing purposes. These data sets were generated by numerically solving the Helmholtz equation for the developed test-case. During the training, these values represent the set of sensor readings that are used to train the network (($u(x_{i},y_{i},z_{i})$) in Figure \ref{fig:CNN}). During the testing, these values represent the reference solution used to assess the accuracy of the network predictions. In this work, we focus on reconstructing the real part of the complex solution of the Helmholtz equation over a 3D domain. These data sets were generated by using Bempp. For reference, see \cite{betcke2019boundary} and \cite{betcke2021bempp}, i.e., a Python based boundary element library for the Galerkin discretization of boundary integral operators in electrostatics, acoustics and electromagnetics in homogeneous bounded and unbounded domains.

\subsection{Network training and spatial field reconstruction}
\label{sec:testing}
\begin{figure}[htb!]
      \centering
      \includegraphics[width=0.45\textwidth]{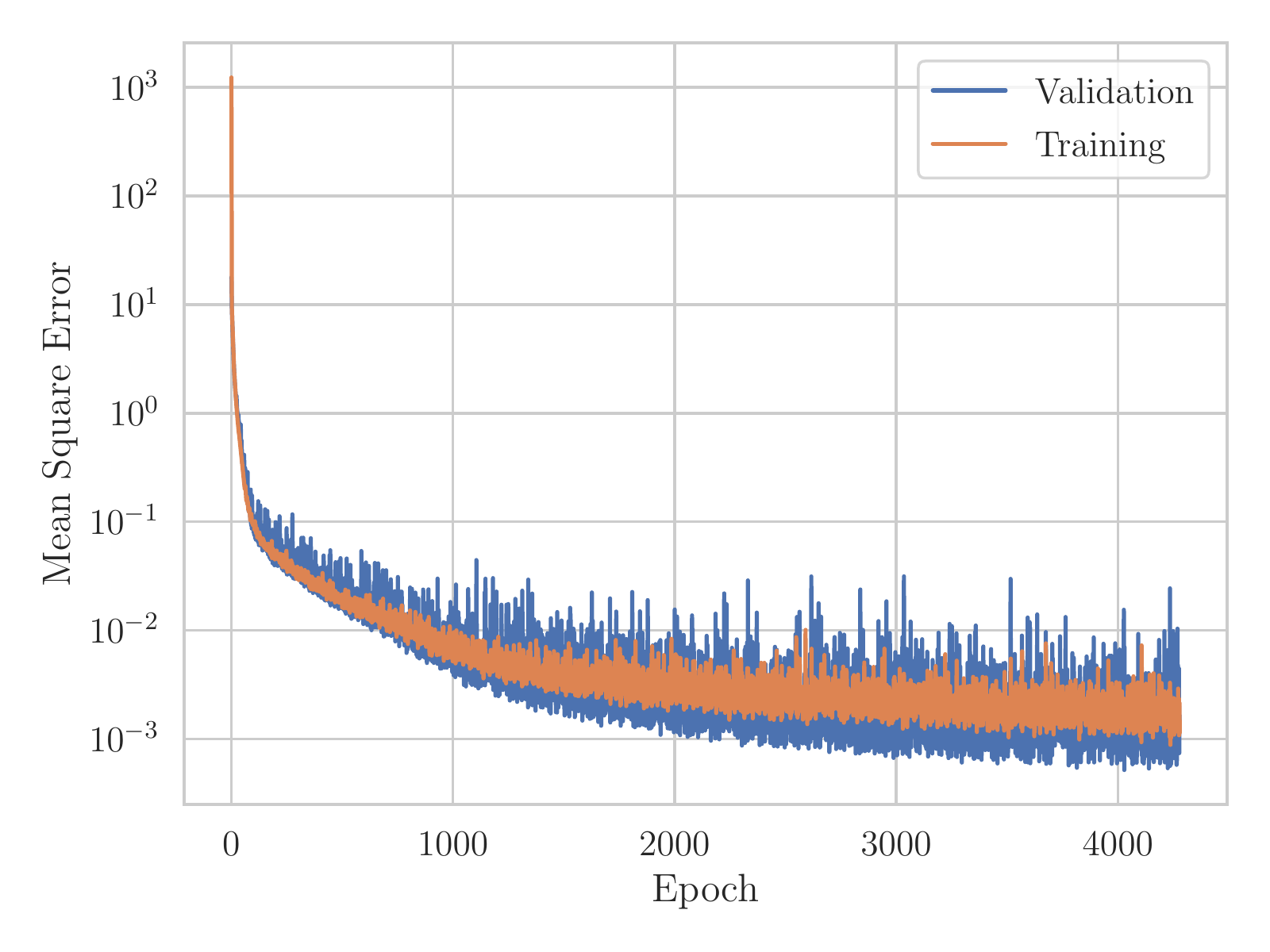}
      \caption{Evolution of the loss function through the epochs during training.}
      \label{fig:trend}
\end{figure}

As mentioned in Section \ref{sec:CNNBVP}, during the training, the convolutional layer outputs the field values at the interior points and, given the Bempp-calculated values in the same locations, the loss function is evaluated and the network is trained. One of the common problems with neural networks is the choice of the number of epochs. Given the limited number of sensor readings inside the domain, the major risk consists in developing an overfitted model. It is common practice to split the original training set into a new training set and a validation set, i.e., a small-size, independent data set that can be used to tune network hyperparameters and deciding when interrupting the training. In particular, the validation data set is constituted by the $20 \%$ of the interior points data set during training.\\
An indication that overﬁtting may have occurred is when the validation error increases whereas the training error still decreases or remains still. In the adopted procedure, a early-stopping criterion is implemented, i.e., an arbitrary large number of training epochs is specified and the training procedure is interrupted when the validation loss stops decreasing. Because of the presence of oscillations that might cause the premature interruption of the training process and the stochastic nature of the optimizer, a delay is added by introducing the “patience” argument ($250$ epochs). Besides, the trajectory of the validation loss was also used to select the best-performing neural network. Figure \ref{fig:trend} depicts the evolution of the losses evaluated during the training of the network. By observing the trajectories, we concluded that the trained network does not suffer from overﬁtting issues, and the model met an optimum solution.

\begin{figure}[htb!]
\begin{subfigure}{.5\textwidth}
\centering
  \includegraphics[width=0.8\linewidth]{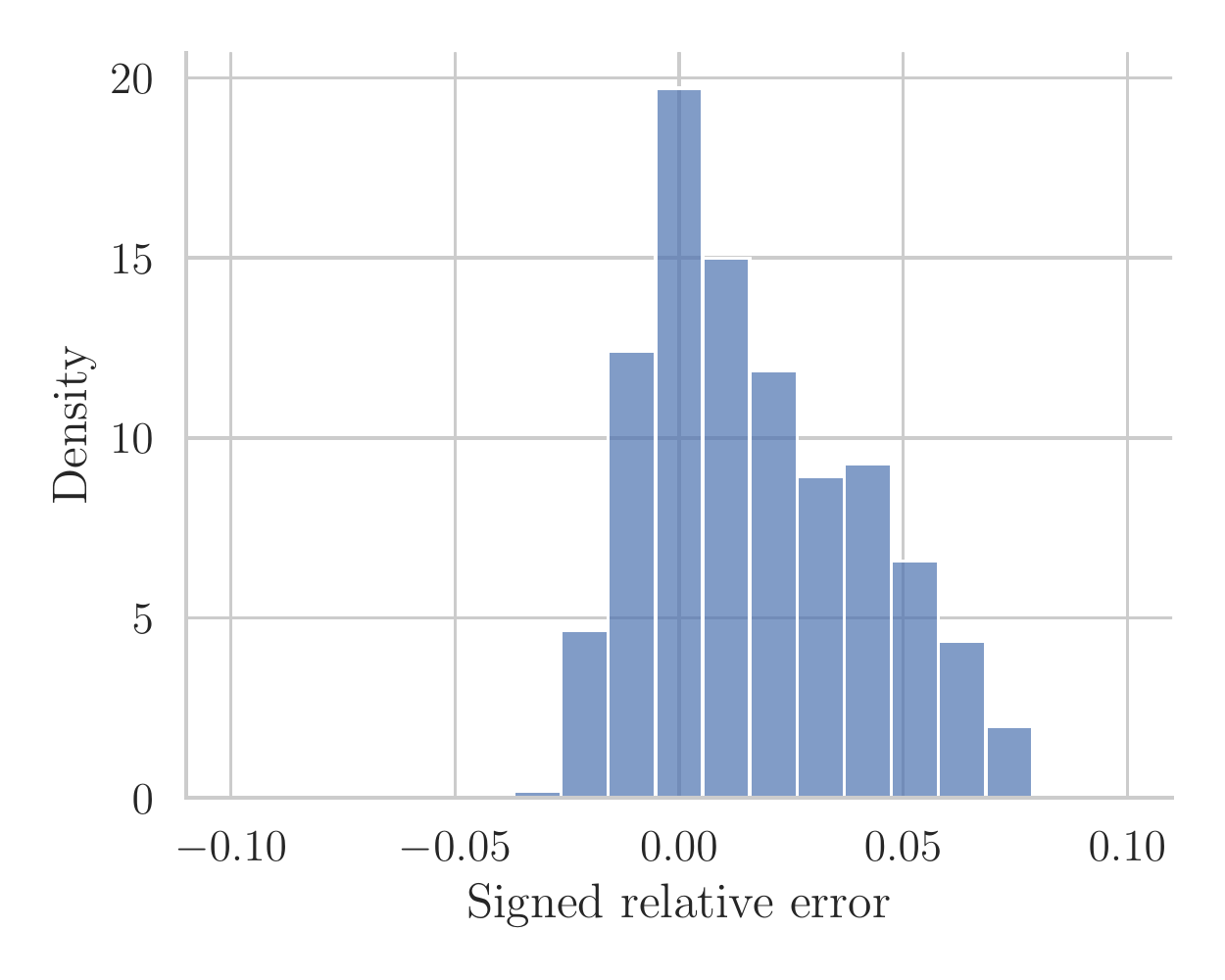}  
  \caption{}
  \label{fig:sgn_error}
\end{subfigure}
\begin{subfigure}{.5\textwidth}
  \centering
  \includegraphics[width=0.9\linewidth]{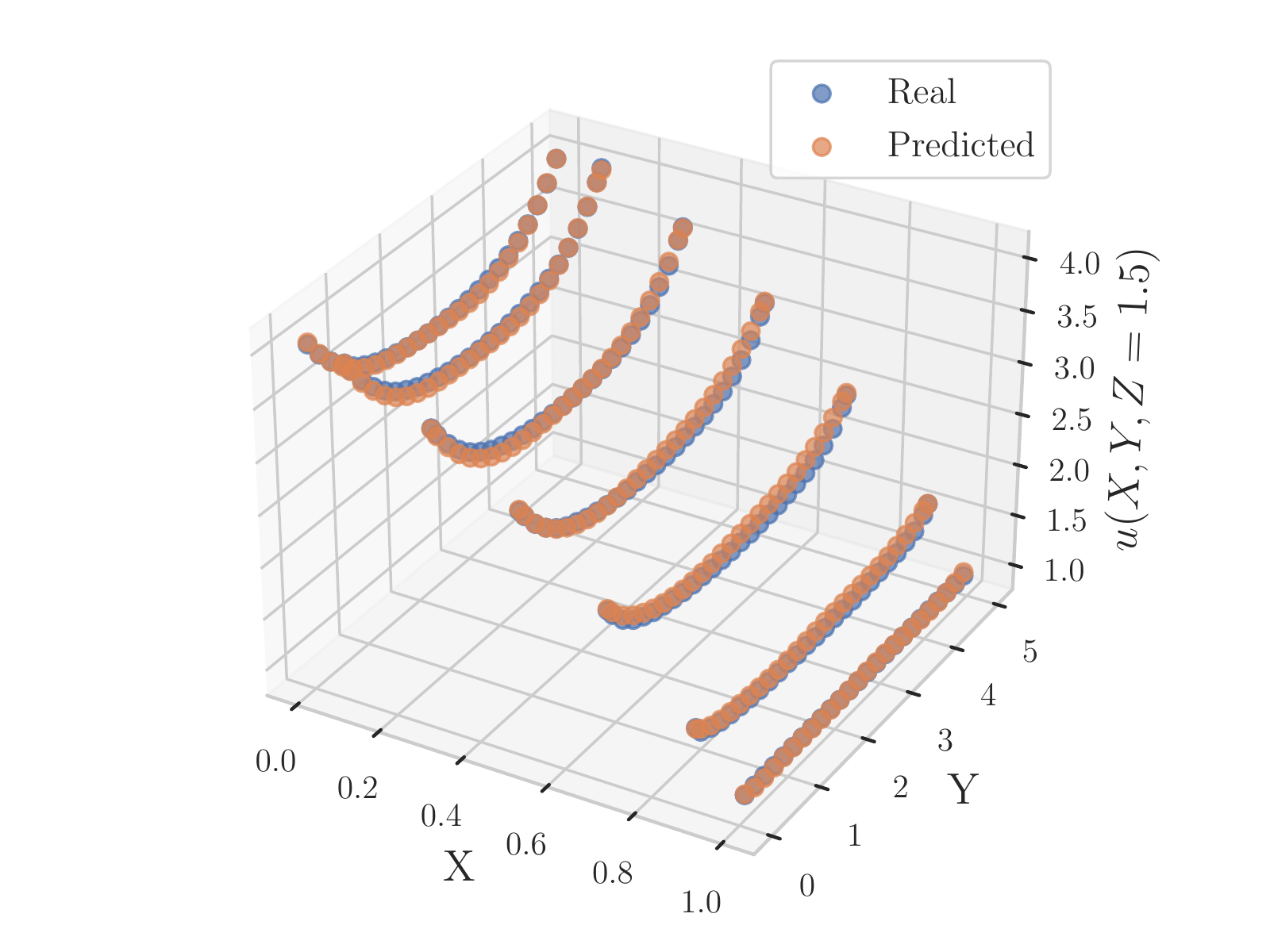}  
  \caption{}
  \label{fig:2D_field}
\end{subfigure}
\caption{Outcomes of the field reconstruction. (a) Probability density function of the signed relative error, (b) comparison between reference solution and network predictions at a certain cross section.}
\label{fig:reconstruction}
\end{figure}

The average wall time required for training and testing the network by using a single worker on a standard laptop (single \SI{3.1}{\giga\hertz} Intel i7 CPU, with 4 cores, and \SI{16}{\gibi\byte} of memory) is \SI{12.71}{\minute}. This wall time refers to the above described test-case, i.e., $15$ interior points within the domain for training/validation, $15,000$ interior points for testing, and $4,600$ collocation points for training/validation and testing. As for the optimization process, maximum $5000$ epochs were adopted, the batch size was set equal to $5$, and the learning rate imposed to Adam optimizer was set equal to $5\cdot10^{-5}$.\\
In Figure \ref{fig:reconstruction}, the outcomes of the field reconstruction are plotted. In Figure \ref{fig:reconstruction}a, the performance all over the 3D domain are represented by showing the probability density function of the signed relative error at the coordinates of the interior points selected for the testing. Results show that most of the predictions are within \SI{5}{\percent} with respect to the Bempp-calculated solution. In Figure \ref{fig:reconstruction}b, the solution evaluated at a certain cross-section of the 3D domain ($z=1.5$) is depicted. The red dots represent the network predictions, the blue dots represent the Bempp-calculated solution.  

\section{Limitations of the CNN-based approach}
\label{sec:method}
\subsection{Sensitivity analyses on network features and physics of the case study}
\begin{figure}[ht!]
\centering
\begin{subfigure}{.45\textwidth}
\centering
  \includegraphics[width=1.0\linewidth]{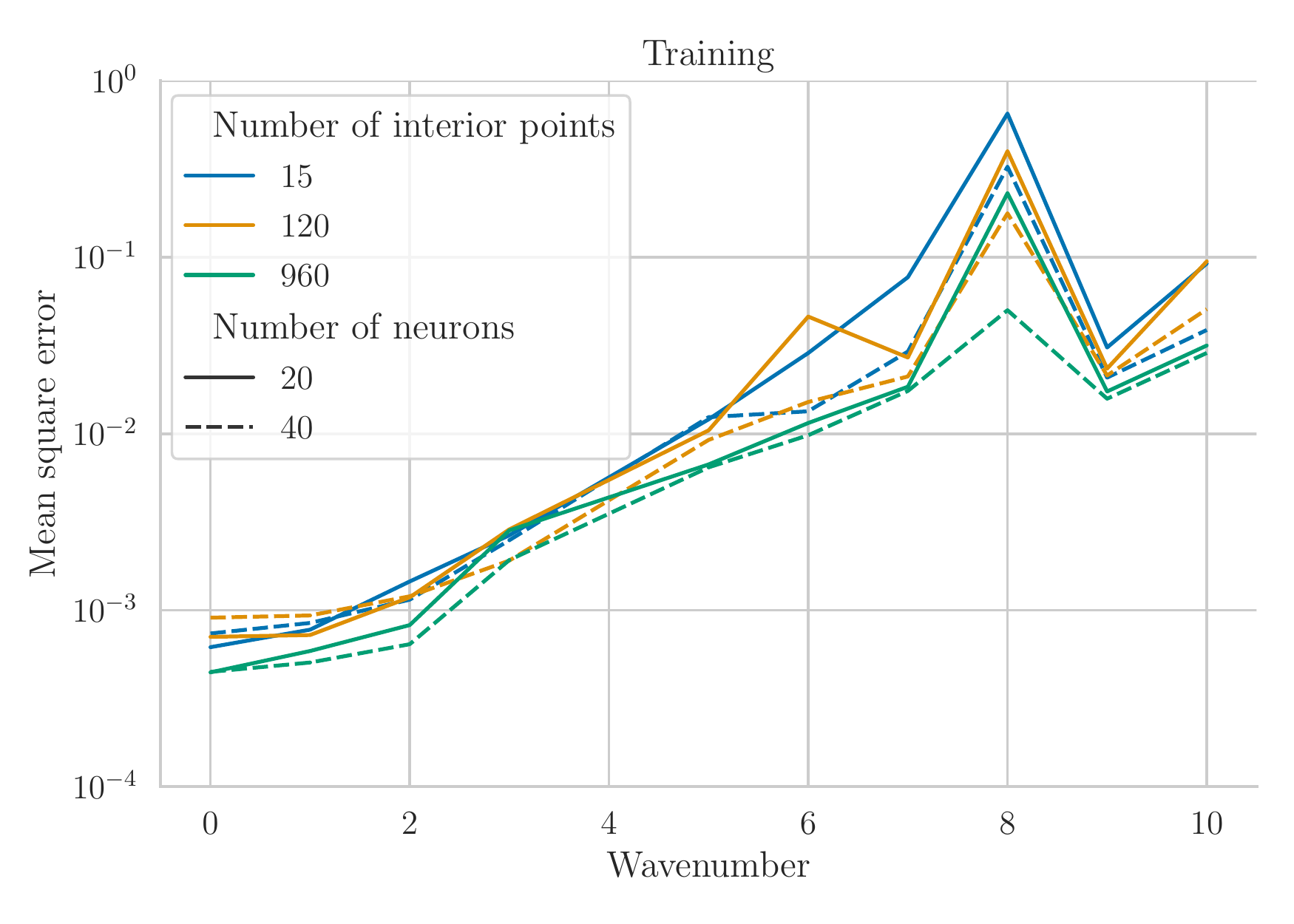}
  \caption{}
\end{subfigure}
\begin{subfigure}{.45\textwidth}
  \centering
  \includegraphics[width=1.0\linewidth]{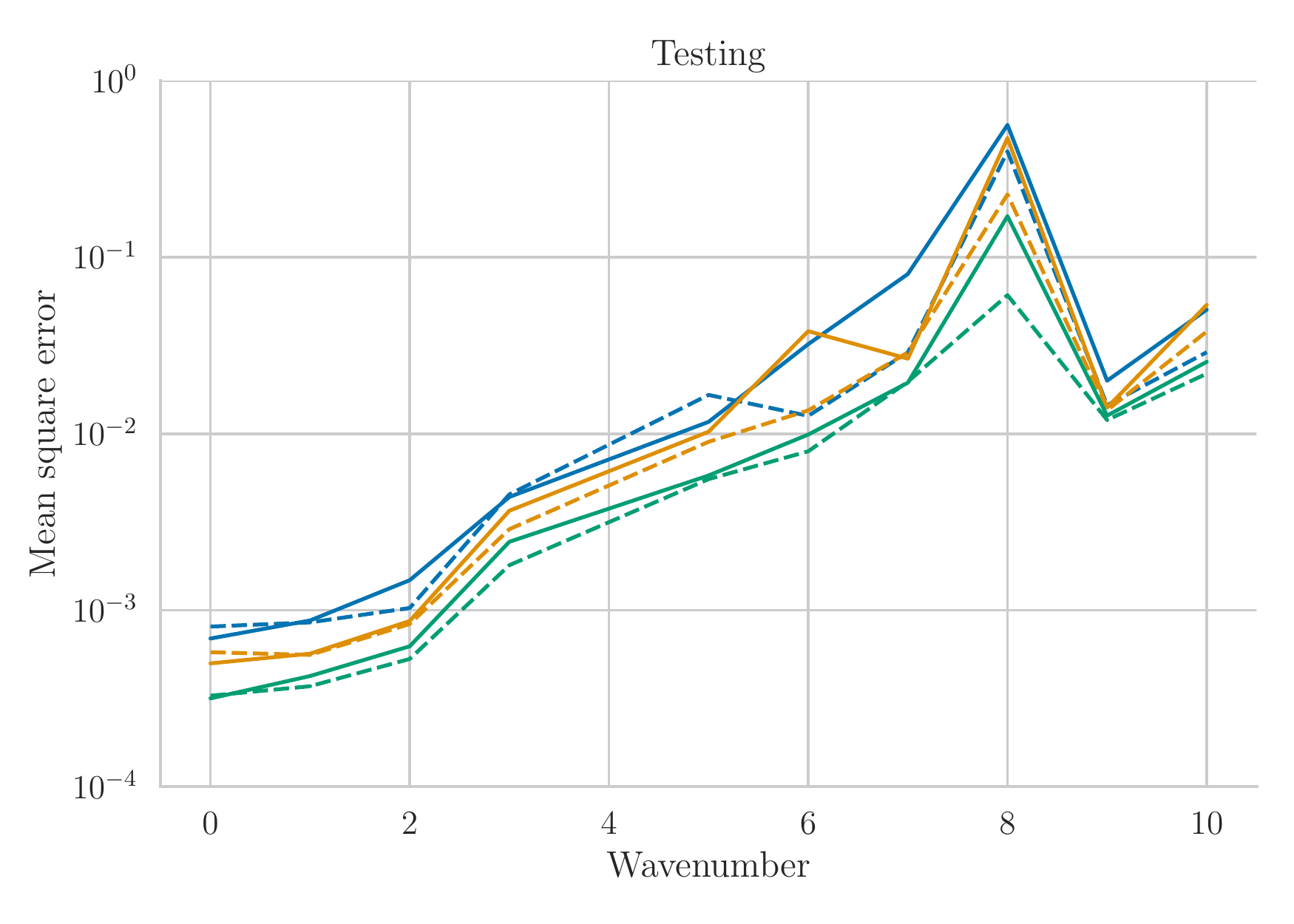}
  \caption{}
\end{subfigure}
\caption{Sensitivity of the network performance as function of the wavenumber, the number of interior points and the number of neurons during (a) training and (b) testing.}
\label{fig:sensitivity}
\end{figure}

To assess the capabilities of the proposed algorithm, a sensitivity analysis was performed by investigating different physical conditions and network configurations. First, conditions with wavenumbers ranging from $0$ to $10$ were examined. All of the calculations were conducted with the same set of boundary conditions (Eqs.(\ref{eq:eq6})(\ref{eq:eq7})), and the same number of collocation points ($4600$ locations, evenly distributed on the domain boundary). Simulations were then repeated by adopting different combinations of the number of interior points and the number of neurons in the dense layers. To evaluate the accuracy of network predictions, the MSE between the network-reconstructed field and the Bempp-calculated solution was used as metrics for each case, as outlined in Section \ref{sec:training}. To limit the possibility of reaching local minima, the networks were trained multiple times by using different random seeds. In Figure \ref{fig:sensitivity}, the value reported at each wavenumber value represents the minimum average loss for the best-performing trained neural network selected over five independent training sessions.

\begin{figure}[ht!]
\begin{subfigure}{.57\textwidth}
\centering
  \includegraphics[width=1.0\linewidth]{figures/Figure_3}
  \caption{}
  \label{fig:intPoints_15}
\end{subfigure}
\begin{subfigure}{.57\textwidth}
  \centering
  \includegraphics[width=1.0\linewidth]{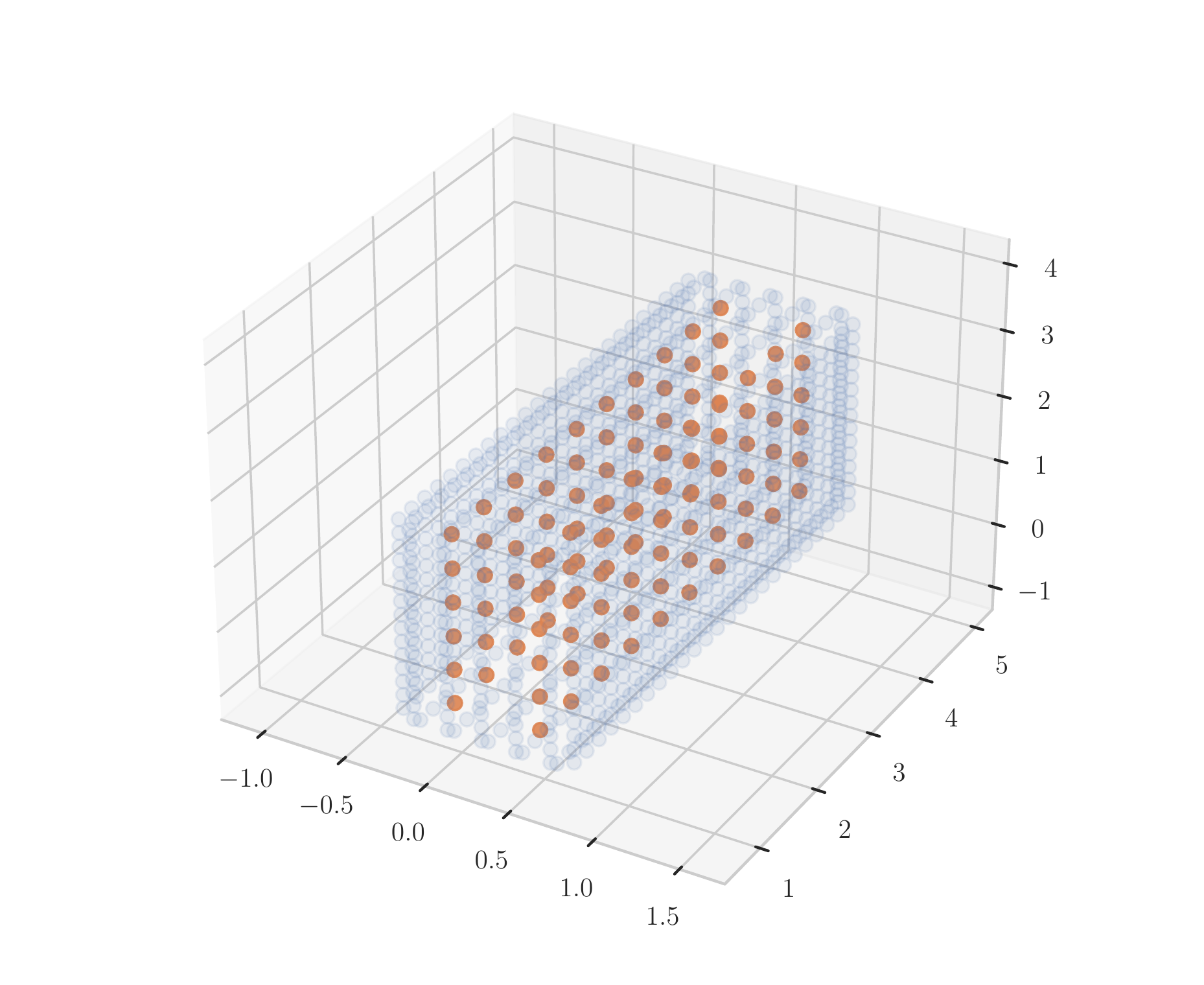}
  \caption{}
  \label{fig:intPoints_120}
\end{subfigure}
\begin{subfigure}{.57\textwidth}
  \centering
  \includegraphics[width=1.0\linewidth]{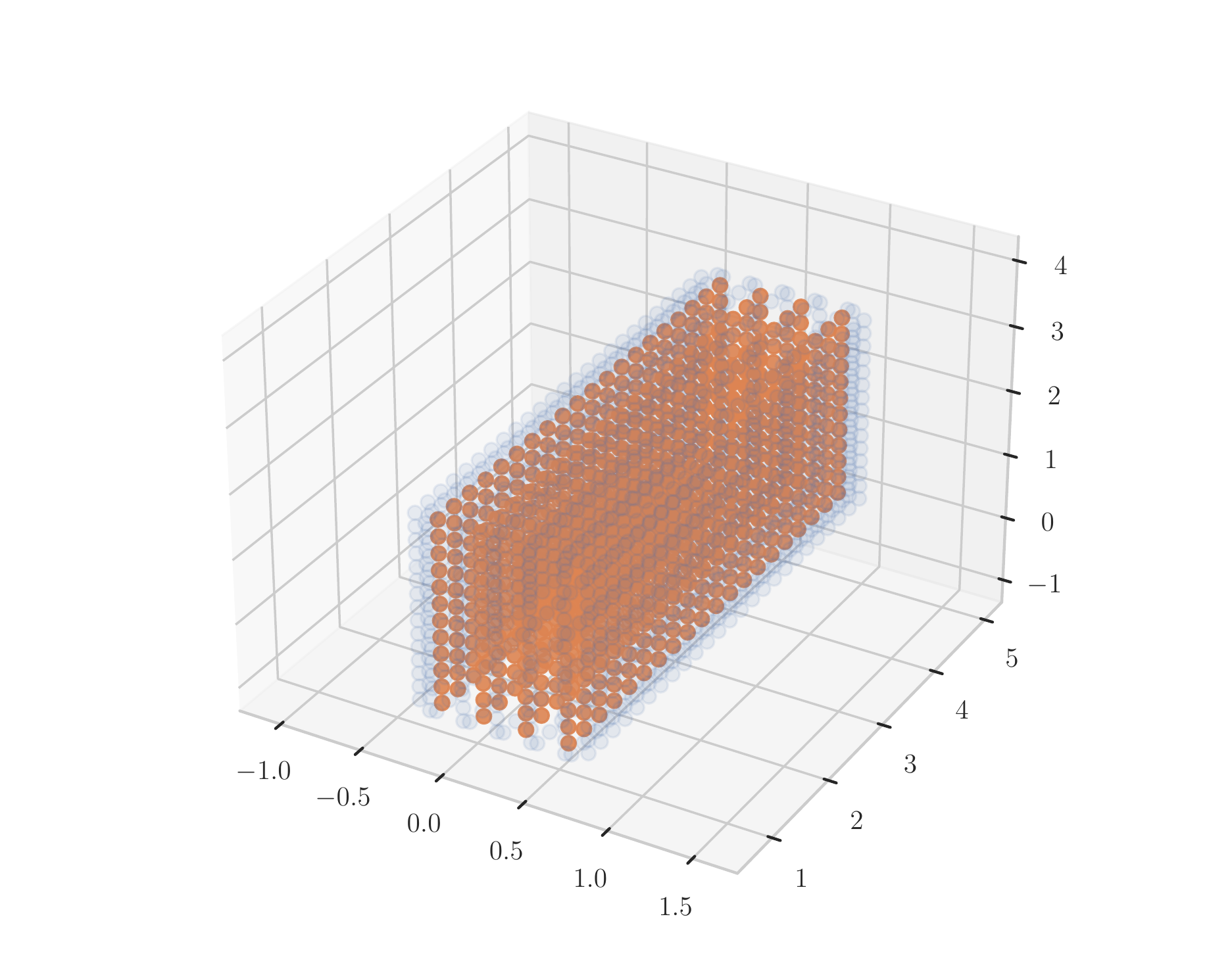}
  \caption{}
  \label{fig:intPoints_960}
\end{subfigure}
\caption{Distribution of the sets of interior points (red dots) and collocation points (blue dots) during the training.}
\label{fig:intPoints_sensitivity}
\end{figure}

\begin{itemize}
\item \textit{Number of interior points}: data sets with different numbers of interior points within the domain were used to train the network. In the sensitivity calculations, $15$, $120$ and $960$ interior points are considered (Figure \ref{fig:intPoints_sensitivity}). These points are uniformly distributed to ensure that each one has the same volume of influence. As a result, three different networks were obtained at the end of the training. During the testing, the three networks were used to reconstruct the field over the same number of interior points ($15,000$ locations).

\item \textit{Number of neurons in the dense layers}: sensitivity calculations were performed by adopting $20$ neuron-layers and $40$ neuron-layers, respectively. The reference network configuration foresees two sets of 3 dense layers each. Simulations were repeated by adopting two sets of 5 layers each, but no sensible improvements were observed (these results are not shown in Figure \ref{fig:sensitivity}). These results proved that two sets of 3 layers with 20 neurons each are sufficient to reconstruct the Green's function and its normal derivative.
\end{itemize}

\subsection{Results interpretation}
The outcomes of the sensitivity analysis shown in Figure \ref{fig:sensitivity} demonstrate that the accuracy of the network predictions are mainly affected by the value of the wavenumber. To confirm the wavenumber degradation dependency, the network performance is further analyzed. From a physical standpoint, a higher wavenumber accentuates the oscillatory nature of the field. This trend can only be represented by using complex numbers. Real parts quantify the amplitude of the oscillations, while the imaginary parts characterize quantities such as phase and frequency. Since this study only considers the real part of the complex field, some aspects cannot be described by the current network design.\\
Despite this limitation, the spatial resolution can be quantified. The set of interior points in the training data set corresponds to the used array of sensors, and their distribution needs to be fine enough so that oscillations can properly be resolved. As shown in Eq.(\ref{eq:error}), the numerical error between the network predictions and the Bempp-calculated solution is bounded by the linear combination of a first-order term, which directly depends on the spatial resolution, and a second-order term that increases with the wavenumber. The linear term ($O(k\Delta \bs{r})$) represents the error due to the domain spatial discretization. The quadratic term ($k O(k\Delta \bs{r})^2$) accounts for the impact of the oscillatory nature of the field.

\begin{equation}
    \epsilon \leq C_1 O(k\Delta \bs{r}) + C_2 k O(k\Delta \bs{r})^2
    \label{eq:error}
\end{equation}

where $C_1$ and $C_2$ are constants that do not depend on the wavenumber or the spatial resolution used in the algorithm.

In Figure \ref{fig:sensitivity}a, the training results are shown.
The MSE monotonically increases with the wavenumber in the range $[0,4]$, but it barely changes with the number of interior points. This trend can be observed where the first-order term in Eq.(\ref{eq:error}) dominates (low wavenumber region). For $k > 4$, the pattern in the loss function is harder to determine for the different networks. The second-order term becomes dominant and the changes in the number of interior points does not have a clear impact on the MSE.\\
In Figure \ref{fig:sensitivity}b, the testing results are shown. The MSE exhibits the same pattern as in the training, i.e., the second-order term becomes dominant for $k > 4$. However, the first-order term shows a more relevant impact in the low wavenumber region. The impact of the number of sensors used in the training is reflected by the distance between the MSE curves, i.e., the MSE corresponding to the network trained with $15$ interior points constitutes the upper bound to the other MSE results. Nonetheless, the accuracy improvements resulting from increasing the number of sensor is overshadowed by the error increase related to the wavenumber. This conclusion is extremely important. It demonstrates that a small number of interior points is sufficient to train a network that ensures accurate predictions. Improving the network performance for higher oscillatory regimes is currently under investigation.

\subsection{Investigation of aliasing in the network predictions}
Domain reconstruction methodologies are limited by the capability of the adopted array of sensors to accurately represent the physical field characteristics. Spatial resolution of these data points ought to be sufficient to not only represent the spatial distribution of the field but also reproduce its dynamic characteristics (i.e., waveform, amplitude and phase). The Nyquist–Shannon sampling theorem outlined in \cite{Shannon} is used to ascertain that the field distribution in the domain is computed with sufficient spatial resolution and that the results do not suffer aliasing. The theorem imposes limits to the maximum spacing between data points, based on the wavenumber, to optimize the reconstruction of the Helmholtz equation solution with the available interior points, i.e., spatial sampling needs to be twice the highest waveform frequency in the domain.

\begin{equation}
k_{sampling} (\bs{r})=2k_{max}(\bs{r})=\frac{2\pi}{\Delta \bs{r}}
\label{eq:eq8}
\end{equation}
where $\Delta \bs{r}=(\Delta x,\Delta y,\Delta z)$.\\
Since the wave propagation characteristics are assumed to be isotropic, aliasing constraints are identical regardless of the direction of propagation.

\begin{equation}
\Delta \bs{r}_{max} = \Delta x_{max} = \Delta y_{max} = \Delta z_{max} = \frac{1}{10} \leq \frac{\pi}{k_{max}}
\label{eq:eq9}
\end{equation}
\begin{equation}
k_{sampling} (\bs{r}) \leq \frac{2\pi}{\Delta \bs{r}_{max}}
\label{eq:eq10}
\end{equation}

The spacing of the collocation and interior points are well within the Nyquist waveform frequency for the sensitivity wavenumbers considered ($k_{max}=10$). Observed variations in the network capabilities to reconstruct the spatial distribution of the field are due to the intrinsic behavior of the neural network and are not due to aliasing.

\section{Conclusion}
\label{sec:conclusion}
In this work, an algorithm to reconstruct the spatial distribution of 3D fields is presented. The presented results demonstrated that the implementation of the BEM within a CNN framework allows reconstructing the Green’s function of the differential operator of a Helmholtz equation, and that it can be used to predict the value of the field in any point of the domain. Here the main features of the proposed scheme are summarized:
\begin{itemize}
\item \textit{Data-driven approach}: the field can be reconstructed from the readings of a discrete array of sensors. The characterization of the physical properties of the monitored domain is not necessary.
\item \textit{Green's function derivation}: thanks to embedding the BEM approach into the neural network, the numerical approximation of the Green's function associated to the differential operator can be obtained for domain of arbitrary geometry. 
\item \textit{Physics-informed, process-agnostic tool}: the only assumption is the applicability of the Kirchhoff-Helmholtz theorem. A wide range of diagnostics problems can be ideally tackled, from the localization of the source of pollutant within a water reservoir to the monitoring of the neutron flux in a nuclear reactor.
\item \textit{Diagnostics applications}: the weights of the network dense layers can be updated during the system evolution. Any deviation with respect to the expected reference conditions can be detected. The developed tool will facilitate the fault diagnostics with a measurement procedure than is less invasive than the current state-of-the-art sensor placement strategies. 
\end{itemize}
Besides the diagnostic applications, the developed tool can be used to support the sensor set design. If the system technical specifications are available, the optimal sensor set (minimum number of sensors ensuring the desired monitoring performance) can be evaluated by performing a sensitivity over multiple distributions of array of sensors. The evaluation of this optimal configuration translates into design, manufacturing, and maintenance cost benefits. In terms of further improvement of the algorithm, the impact of experimental uncertainties affecting the data still needs to be evaluated. In this work, training data-sets have been obtained by adopting simulation outcomes. The adoption of real data and the benchmark of the obtained predictions will confirm the performance of the developed algorithm.

\bibliographystyle{unsrtnat}
\bibliography{references}  

\section*{Acknowledgements} \label{sec:ackw}
This material is based upon work supported by the U.S. Department of Energy, Office of Science, under contract DE-AC02-06CH11357.

\end{document}